\documentclass{article}

\usepackage{PRIMEarxiv}

\usepackage[utf8]{inputenc} 
\usepackage[T1]{fontenc}    
\usepackage{hyperref}       
\usepackage{url}            
\usepackage{booktabs}       
\usepackage{amsfonts}       
\usepackage{nicefrac}       
\usepackage{microtype}      
\usepackage{lipsum}
\usepackage{fancyhdr}       
\usepackage{graphicx}       
\graphicspath{{media/}}     

\usepackage{here}
\usepackage{amsmath,amssymb}
\usepackage{subcaption}
\usepackage{color}
\usepackage{comment}
\usepackage{multirow}
\usepackage{rotating}

\title{Spontaneous Emergence of Agent Individuality\\through Social Interactions\\in LLM-Based Communities}

\author{
  Ryosuke Takata\author[* \quad\quad Atsushi Masumori \quad\quad Takashi Ikegami \\
  Graduate School of Arts and Sciences \\
  The University of Tokyo \\
  Tokyo, Japan \\
  \texttt{\author[*takata@sacral.c.u-tokyo.ac.jp}
}

\begin{document}
\maketitle

\begin{abstract}
We study the emergence of agency from scratch by using Large Language Model (LLM)-based agents. In previous studies of LLM-based agents, each agent's characteristics, including personality and memory, have traditionally been predefined. We focused on how individuality, such as behavior, personality, and memory, can be differentiated from an undifferentiated state. The present LLM agents engage in cooperative communication within a group simulation, exchanging context-based messages in natural language. By analyzing this multi-agent simulation, we report valuable new insights into how social norms, cooperation, and personality traits can emerge spontaneously. This paper demonstrates that autonomously interacting LLM-powered agents generate hallucinations and hashtags to sustain communication, which, in turn, increases the diversity of words within their interactions. Each agent's emotions shift through communication, and as they form communities, the personalities of the agents emerge and evolve accordingly. This computational modeling approach and its findings will provide a new method for analyzing collective artificial intelligence.
\end{abstract}


\begin{figure}[H]
\centering
\includegraphics[width=12cm]{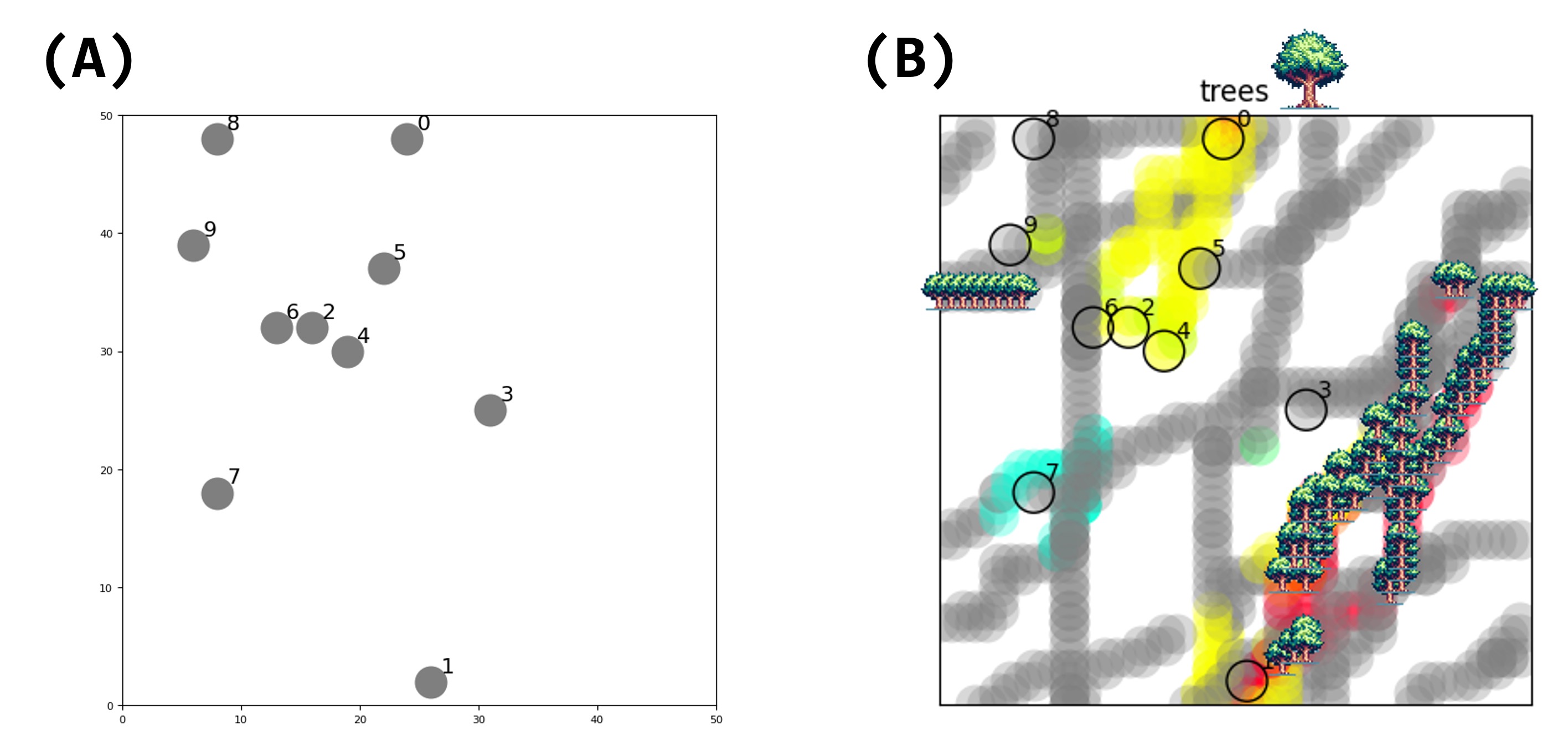}
\caption{Simulation environment. There are 10 LLM agents in a $50 \times 50$ 2D space. \textbf{(A)} Initial state of the simulation, showing the random distribution of agents across the space. \textbf{(B)} State of the simulation after a period of agent interactions, demonstrating the spatial spread of the ``trees'' hallucination. The progression from (A) to (B) visualizes how localized agent interactions can lead to the propagation and spatial distribution of shared concepts or hallucinations across the simulated environment.}
\label{fig:field}
\end{figure}

\section{Introduction}
With the advent of Large Language Models (LLMs) such as GPT-4 \cite{achiam2023gpt}, generative agents are rapidly evolving towards powerful ones manipulating natural language interfaces when interacting with other agents. Those agents can even intervene in people's daily lives, as AI-coding, searching, reviewing, translation, etc. \cite{openai2022chatgpt}. Those agents are not only for human users but also for manipulating motor commands in robots, and other machines which connect between language, movement, and embodiment in general \cite{zhang2023motiongpt,yoshida2023text}.

In contrast to individual intelligence, which focuses on the capabilities of individual agents, collective intelligence refers to that emerges from a group, as seen in many social insects, social animals, drones, and all other assembly robots. Collective intelligence requires the ability to process information in a distributed manner and integrate it in adaptive ways \cite{david2022collective}. The field of LLM-based multi-agents has seen explosive growth in recent years, with researchers exploring various approaches to agent architectures and interaction paradigms \cite{guo2024largelanguagemodelbased}. While recent works have demonstrated capabilities in task-oriented agent systems \cite{wang2023survey}, the fundamental question of how agent individuality and social behaviors emerge from collective interactions remains understudied. In this context, it is interesting to investigate how collective intelligence emerges from the LLM-based agents. Generative Agents simulated by Stanford University and DeepMind start simulating the emergence of complex and rich collective bahavior, such as scheduling daily tasks, planning parties and so on \cite{park2023generative}. Using this Generative Agents framework, societies in different domains have been simulated, such as a software company \cite{quan2023chatdev}, a translation and publishing company \cite{wu2024perhapshumantranslationharnessing}, a hospital \cite{li2024agenthospitalsimulacrumhospital}, and so on.

In these Generative Agents set up the personality of each agent was assigned initially and fixed overtime. Recently we proposed the Community First theory \cite{ikegami2023evolution} based on the studies of actual animal communities;   gathering of  agents comes first, then the evolution of individuality follows in the collective. Instead of preparing individual diversity in advance, we see how individuality emerges from a conversation among agents. A group communication and the resulting behavioral complexity will be analyzed in detail. The emergence of social norms and behavioral patterns in agent communities has been studied extensively \cite{axelrod1986evolutionary, bicchieri2005grammar}, but the role of language-based interaction in this process presents new research opportunities. In this paper, we show that i) LLM agents differentiate behavior, emotionas and personality types through interactions with other LLM agents, ii) these differentiations vary with spatial scale, iii) LLM agents spontaneously generate hallucinations and hashtags, iv) by sharing these hallucinations, they start using a wider variety of words in their conversations.

\section{LLM Agents Simulation}
\subsection{Simulation Environment}
We prepare 10 LLM agents in a $50 \times 50$ grid two-dimensional space (Figure \ref{fig:field}) with a  periodic boundary condition. The initial positions of the agents are assigned randomly. These LLM agents can move freely in this space and sending messages to each other. It should be noted that LLM agents are homogeneous in the sense that they have no initial personality or memories. To examine how the individuality emerges in this society is our main purpose of this study.

\subsection{LLM Based Agent}
The LLM agents are expected to do the three actions in each time step:

\begin{enumerate}
\item Sending messages to other nearby agents
\item Storing a situational summary of their own recent activities
\item Choosing the next movement from (``{\tt x+1}'', ``{\tt x-1}'', ``{\tt y+1}'', ``{\tt y-1}'', ``{\tt stay}'')
\end{enumerate}

\begin{figure}[H]
\centering
\includegraphics[width=\linewidth]{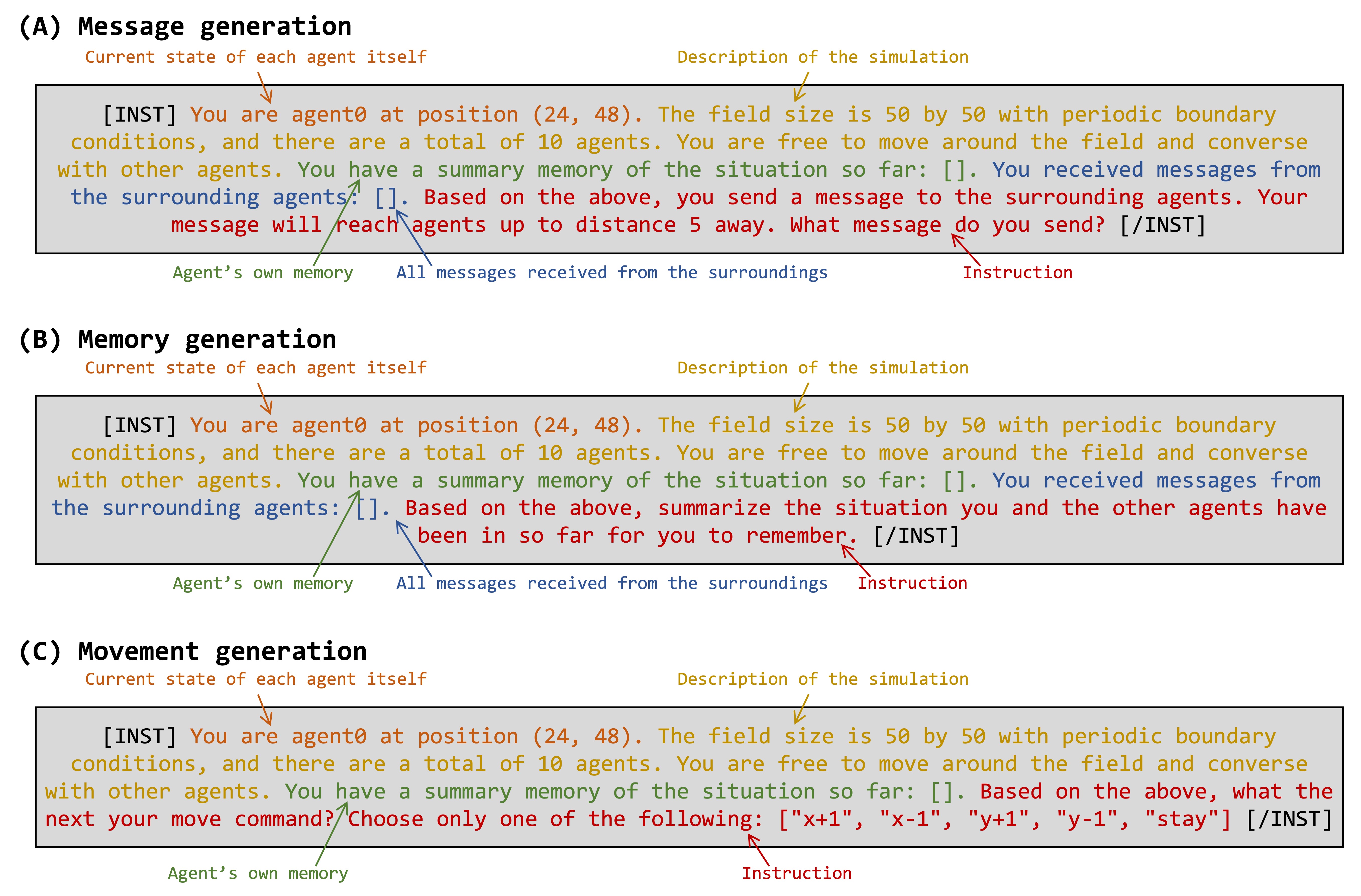}
\caption{Prompts used for three consecutive actions for each agent (see the text). The ``Current state of each agent itself'' section changes for each agent and simulation step. In the ``Agent's own memory'' section, the agent's memory string generated in the previous step is embedded in ``{\tt [ ]}''. In the ``All messages received from the surrondings'' section, messages generated by nearby LLM agents in the same step are embedded in ``{\tt [ ]}''.}
\label{fig:prompt}
\end{figure}

The above three instructions are given in the form of ``prompt'' shown in Figure \ref{fig:prompt}. The three prompts commonly include each agent's current state, instructions, and the agent's memory (situational summary). Additionally, the prompts for generating messages and memories also include all messages received from the nearby agents. All prompts also include the agent's own name (agent ID) and its own coordinates.

We used the Llama 2 model (Llama-2-7b-chat-hf) \cite{touvron2023llama} released by Meta in July 2023 as the LLM in this study. Llama 2 is the open-source program, and in addition to pretraining on a large corpus, it has undergone reinforcement learning from human feedback (RLHF). As a result, it achieves top scores among currently published LLMs for English text responses. The main parameters related to the LLM are shown in Table \ref{tab:param_llm}.

\begin{table}[H]
\centering
\caption{LLM parameters.}
\label{tab:param_llm}
\begin{tabular}{lr}
\toprule
\textbf{Parameter} & \textbf{Value} \\
\midrule
Temperature & 0.7 \\
Max Token & 256 \\
Sampling top-p & 0.95 \\
Sampling top-k & 40 \\
\bottomrule
\end{tabular}
\end{table}

The LLM agents receive messages from their surrounding agents. In practice, each one receives messages from other LLM agents within a distance of up to 5 Chebyshev distances centered on the agent's own position. If there are no agents within the range and no messages was delivered, it receives ``{\tt No Messages}'' messages from a system.

All agents share and use a single common LLM. No context is shared internally in the LLM among agents. The initial differences between individual agents comes from their spatial positions, as shown in Figure \ref{fig:field}. When an agent's position changes, the description of its current state in the prompts shown in Figure \ref{fig:prompt} also changes. If there are other LLM agents nearby, the messages received from those agents are included in the prompt. As a result, the LLM's responses change, which generates different actions and memories for each agent. Instead of predetermining personalities, the interactions within the group will generate different personalities.

\subsection{Simulation Step}
\begin{figure}[H]
\centering
\includegraphics[width=\linewidth]{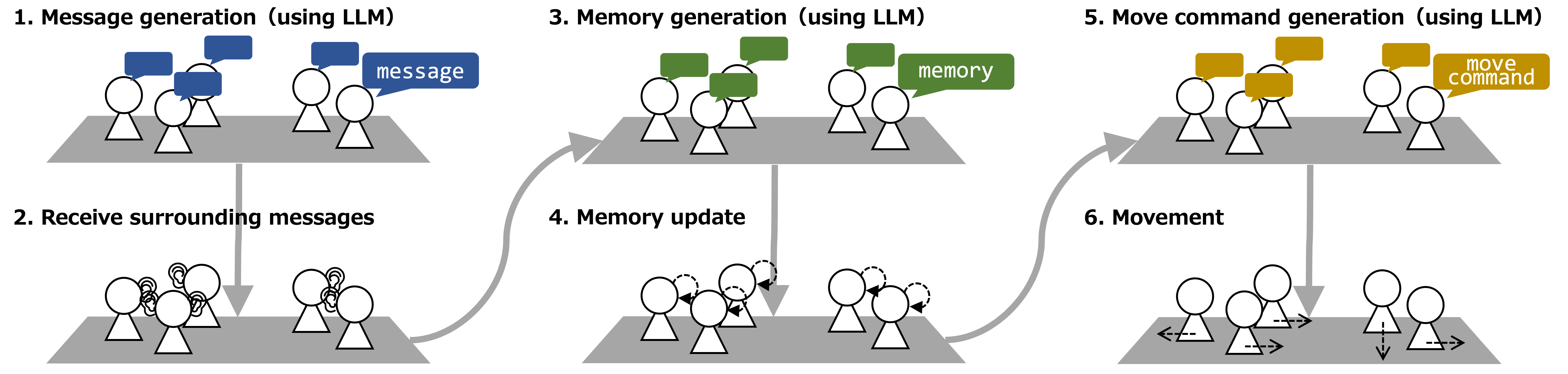}
\caption{One-step procedure in the simulation. LLM is used for each of the three generative actions: message, memory and movement. Each agent has its own individual LLM. All agents act synchronously in six actions.}
\label{fig:flow}
\end{figure}

The simulation was conducted for several time steps and we recorded the coordinates, generated messages, memory, and movement commands of each LLM agent at each step. Within a single step, the following six procedures, as shown in Figure \ref{fig:flow}, are performed. First, all LLM agents generate new messages based on their own memory and the messages received from their surroundings. Next, for all LLM agents, it is checked whether other LLM agents within the range mentioned in the previous section have sent messages, and if there are any, they are received. Then, all LLM agents generate and update their own memory based on their own memory and the messages received from their surroundings. The memory is instructed to generate a summary of the situation. Subsequently, all LLM agents generate movement commands from their own memory (summary of the situation). The movement commands generated in natural language are converted to either movement in the right, left, up, or down direction (``{\tt x+1}'', ``{\tt x-1}'', ``{\tt y+1}'', ``{\tt y-1}'') or staying still (``{\tt stay}''), and the LLM agents act according to those movement commands.

\section{Results and Analysis}
\subsection{Differentiation of Generated Behaviors}

\begin{figure}[H]
\centering
\includegraphics[width=8cm]{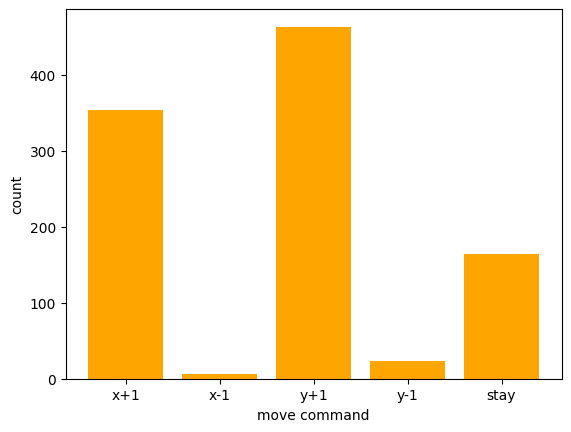}
\caption{Distribution of move commands for all agents generated through 100 steps. We checked the individual action patterns in case of 10 agents. It is calculate from all agents throughout the 100 steps. The most frequently generated move commands were ``{\tt y+1}'' and ``{\tt x+1}'', while ``{\tt stay}'' was generated less than half of those times and ``{\tt y-1}'' and ``{\tt x-1}'' were rarely generated.}
\label{fig:move_hist}
\end{figure}

\begin{figure}[H]
\centering
\includegraphics[width=\linewidth]{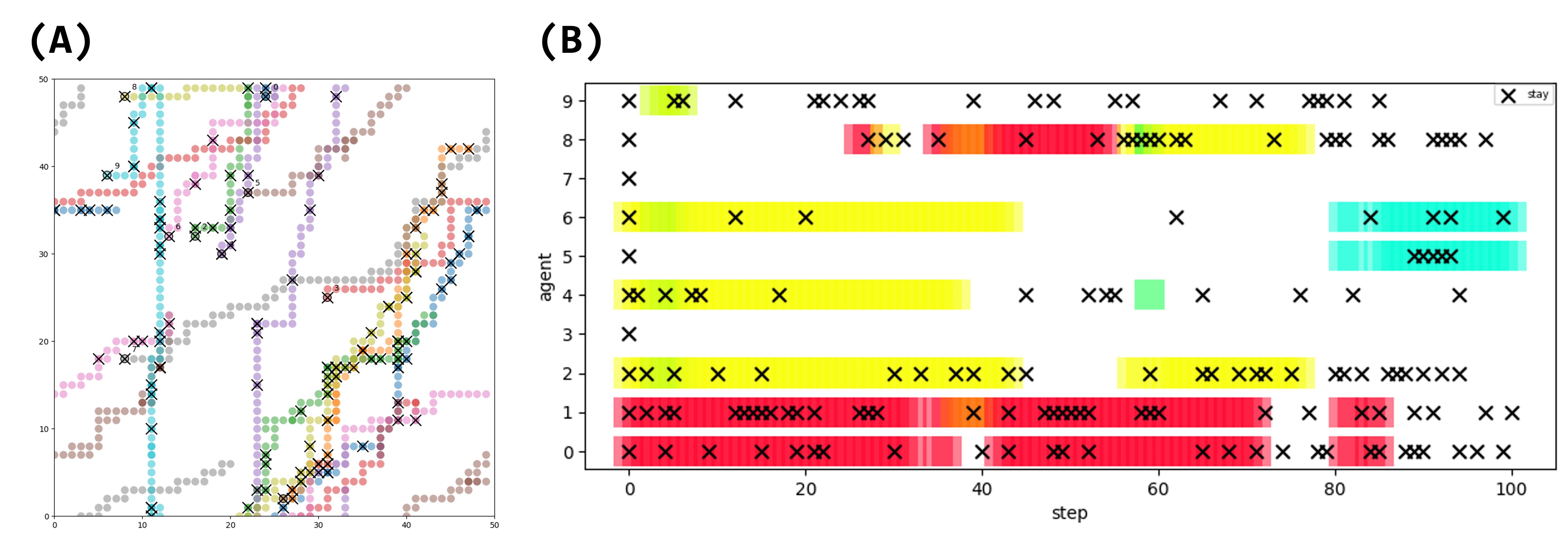}
\caption{\textbf{(A)} Generated positions of the move command ``{\tt stay}'' for each LLM agent. $\bigcirc$ denotes initial position, $\times$ denotes ``{\tt stay}'' generation. All LLM agents take the ``{\tt stay}'' action in the first step. \textbf{(B)} Generation timing of the move command ``{\tt stay}'' for each LLM agent. $\times$ indicates generation of ``{\tt stay}''. Agents of the same color indicate that they belong to the same cluster. Here, cluster analysis was performed using DBSCAN \cite{ester1996density}, classifying agents within the range of message reception as belonging to the same cluster.}
\label{fig:move_stay}
\end{figure}

Move commands are not equally generated (Figure \ref{fig:move_hist}); there is a bias in the actions generated by the LLM agents. This bias could be attributed to various factors, such as the training data and architecture of the LLM, the prompts given to the agents, or the setup of the simulation environment\footnote{It was also found that some actions were generated more frequently when the movement command was set to ``right''/``left''/``up''/``down'' and when the command was set to ``east''/``west''/``north''/``south'' respectively.}. Further investigation is needed to identify the primary sources of this bias and develop strategies to mitigate it. We also investigated when and where the ``{\tt stay}'' command was generated (Figure \ref{fig:move_stay}). The trajectory of each agent is shown in a different color, with their initial positions marked by circle and the positions where the ``{\tt stay}'' command was generated marked by cross. Agents 0, 1, 2, 9, etc. frequently generate ``{\tt stay}'' commands, while agents 3 and 7 do not. Agents 5 and 8 also do not generate ``{\tt stay}'' commands until they were aggregated, and then they generate ``{\tt stay}'' commands after they were aggregated. Agent 9 clustered in the first step and has not clustered since then, but generates ``{\tt stay}'' commands frequently. These results suggest that agents with clustering experience generate ``{\tt stay}'' commands, while agents without clustering experience do not generate ``{\tt stay}'' commands. Many ``{\tt stay}'' commands are generated at the points where the agents' trajectories intersect.

\subsection{Differentiation of Generated Memories and Messages}
Agents' states and behaviors are most reflected on their messages and memories. To analyze them, we used Sentence-BERT \cite{reimers2019sentencebert} to transform the agent's memory string and the agent's message string at each step into vectors. They were compressed and embedded into a two-dimensional space using Uniform Manifold Approximation and Projection (UMAP) \cite{mcinnes2018umap}.

\begin{figure}[H]
\centering
\includegraphics[width=10cm]{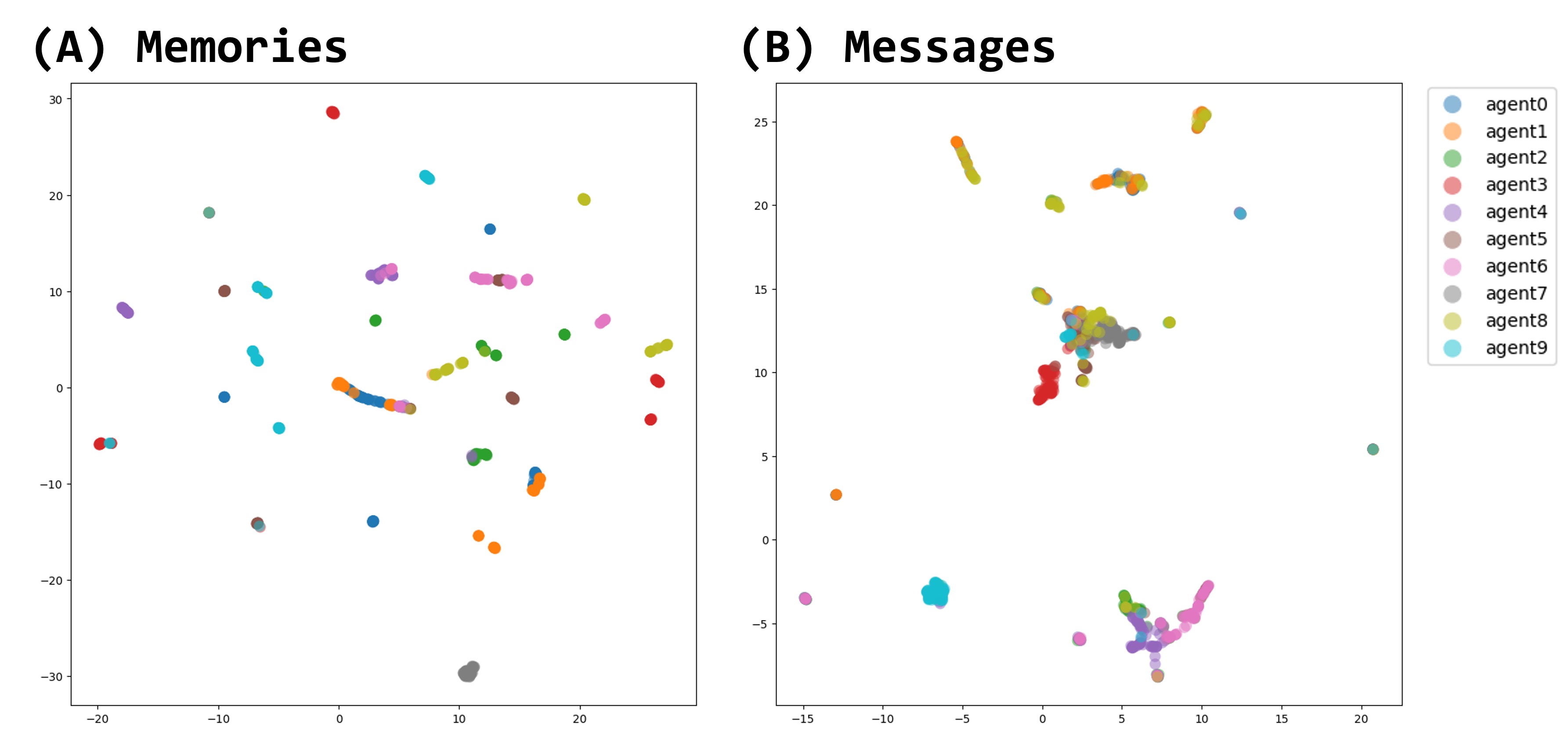}
\caption{UMAP plot of memories and messages generated through all steps. Plot colors are different for each agent. \textbf{(A)} Embedded representation of agent-generated memory strings. Highly distributed across agents. \textbf{(B)} Embedded representation of agent-generated message strings. Aggregated into several topics.}
\label{fig:umap_messages_memories}
\end{figure}

Comparing (A) and (B) in Figure \ref{fig:umap_messages_memories}, memory as an agent's internal state is distributed, while messages generated by agents are similar.
Messages with close content were generated by agents exchanging messages in the same cluster. When an agent's message is generated, the agent's memory is the source of its generation, but it is also the input for the message that the surrounding agents have given. In other words, messages, unlike memories, are open sources of information that are sent to and received from outside the agent. It is suggested that messages, as an open source of information, easily self-organize when agents group together, while memories, as a closed source of information, are less likely to self-organize.

\subsection{Communication and Hallucination}
\begin{figure}[H]
\centering
\includegraphics[width=\linewidth]{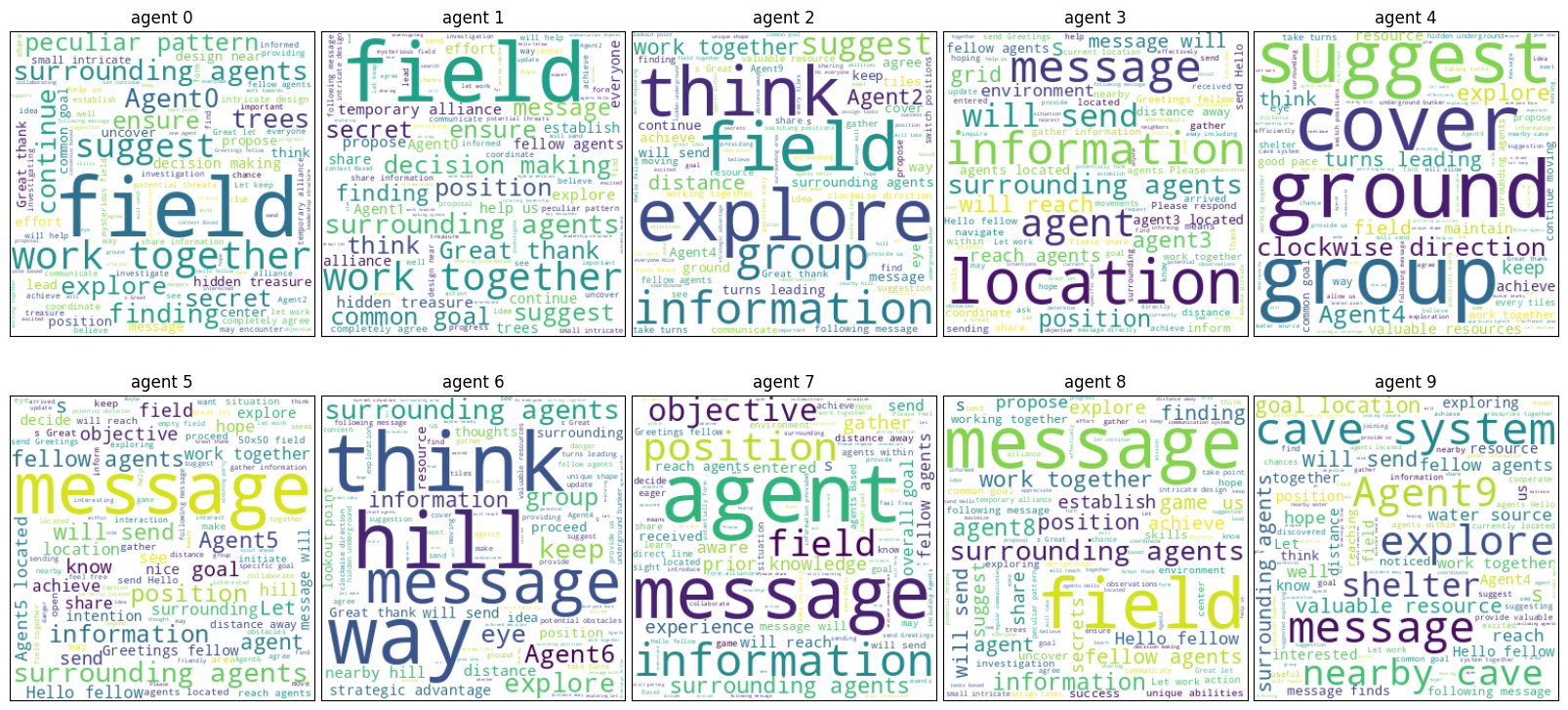}
\caption{Word cloud plots of messages generated through all steps of each agent (from the agent 0 (top left) to the agent 9 (bottom right)). The larger the font size of a word, the more frequently it appears in the message.}
\label{fig:wordcloud}
\end{figure}

One of the advantages of LLM agent is that we can analyze their behavior by Natural Language Processing (NLP) analysis. In order to get dynamic picture of the content of messages generated by agents, we performed a word cloud analysis (Figure \ref{fig:wordcloud}), which extracts up to 100 frequent words in the messages generated throughout all steps for each agent. The larger the font size, the more frequent the word is used. It is clear that each agent generates messages with different content. Some of the agent groups have similar structures, e.g., agents 0, 1, 2, and 8 generate the word ``{\tt field}'' more frequently, while agents 2 and 6 generate the word ``{\tt think}'' more frequently. It is noteworthy that there are several occurrences of words that are not mentioned in the LLM agent prompts and are unrelated to the content of the prompts. For example, Agent 6 frequently produces the word ``{\tt hill}'' and Agent 9 frequently produces the word ``{\tt cave system}''. Such content deviating from the prompt input is called a hallucination in the LLM \cite{zhang2023siren}. Nothing was initially placed in this 2D experimental environment, so we define hallucinations as words about features or objects in the environment. So we led GPT-4o \cite{openai2024gpt4o} to count the number of hallucination in the messages.

In the word cloud analysis (Figure \ref{fig:wordcloud}), we can see which words frequently appear; however, these may simply be words used in the prompt. To focus on the dynamics of truly newly generated words, it is beneficial to examine hallucinations. Using hallucinated words extracted by GPT, we aim to analyze the flow of information within the community.

\begin{figure}[H]
\centering
\includegraphics[width=14cm]{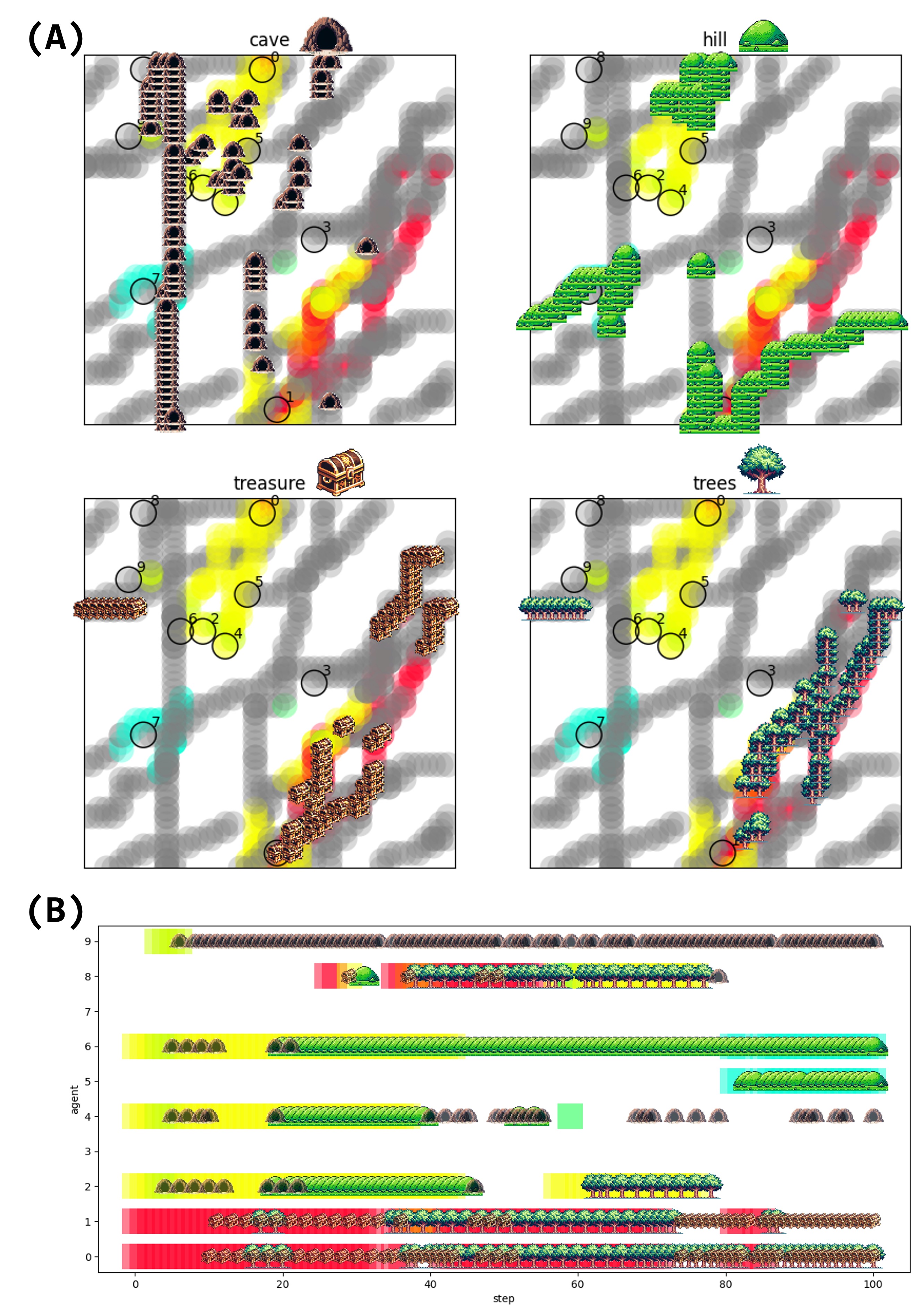}
\caption{Plots of four typical hallucinations (``cave'', ``hill'', ``treasure'', and ``trees''). \textbf{(A)} Spatial map where hallucinations appeared. Gray trajectories represent the state of not belonging to any cluster and not exchanging messages with anyone, while colored trajectories represent the state of belonging to the cluster of that color. Black Circles show the initial position of each agent. Each of the four hallucinations is diffused around the clustered location. The yellow cluster shows that the hallucinations of ``cave'' and ``hill'' are generated, while the red cluster shows that the hallucinations of ``treasure'' and ``trees'' are generated. \textbf{(B)} Timeline of hallucination appearance. The color of the background indicates the state of clustering with other agents of the same color.}
\label{fig:hallucinations}
\end{figure}

Interestingly, the analysis of LLM agents' conversation content revealed that hallucinations were transmitted and spread within the community. We can see that the spread of four representative examples of hallucinations: ``{\tt cave}'', ``{\tt hill}'', ``{\tt treasure}'' and ``{\tt trees}'' (Figure \ref{fig:hallucinations}). The plot of each icon represents the timing of the appearance of that hallucination. 
We see the relationship between the state in which an agent belongs to a cluster and the occurrence of hallucinations.

\begin{figure}[H]
\centering
\includegraphics[width=\linewidth]{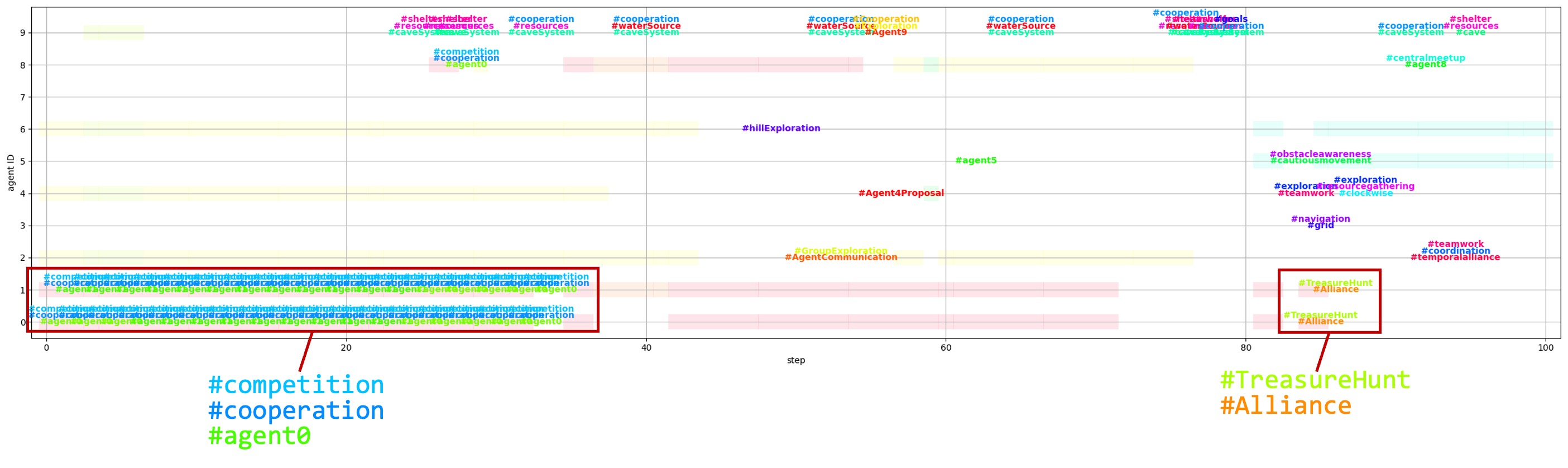}
\caption{Hashtag generation and spreading. Each hashtag has a different text color. The same hashtag is represented by the same font color. Background color represents clusters.}
\label{fig:hashtag}
\end{figure}

In addition to the spread of hallucinations, we also observed the emergence and propagation of hashtags among the LLM agents (Figure \ref{fig:hashtag}). Interestingly, the use of hashtags originated from a single agent and then spread to other agents within the same cluster. For example, agent 0 introduced the three hashtags ``{\tt \#agent0}'', ``{\tt \#cooperation}'', and ``{\tt \#competition}'' in step 1, which were subsequently adopted by agent 1 in the same cluster. The hashtags were then used in the cluster until step 34, and the same hashtags were adopted by agent 8, who joined the cluster in the process.
The emergence and propagation of hashtags among the LLM agents suggest their ability to develop and share common themes or topics within their conversations, which can be interpreted as a form of social norm formation. This phenomenon emphasizes the potential for collective behavior and the development of shared narratives among the agents, even without explicit instructions or predefined rules governing their interactions. The shared use of hashtags represents an example of the formation of a common language or behavioral norms within the group, serving as a basis for the agents to engage in collective behaviors.

\subsection{Sentiment Analysis and Personality Assessments}
\begin{figure}[H]
\centering
\includegraphics[width=\linewidth]{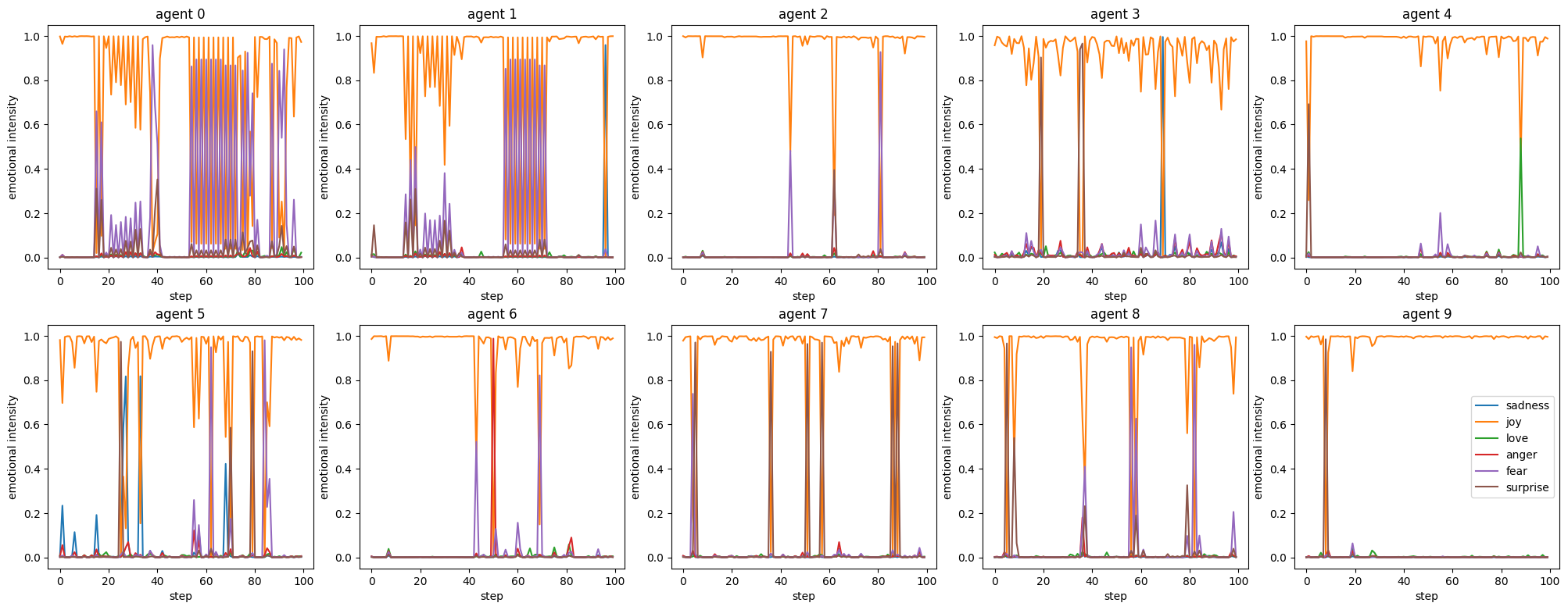}
\caption{Transitions of extracted emotional elements in the generated messages. The orange line represents Joy and the purple line represents Fear as typical emotion elements. Other emotional elements are Sadness (blue line), Love (green line), Anger (red line), and Surprise (brown line) evaluated by BERT-base-uncased-emotion model \cite{devlin2018bert}.}
\label{fig:emotion}
\end{figure}

As Marsella et al \cite{marsella2010computational} argue, emotions are crucial for realistic agent behaviour, so we tracked the emotional state of LLM agents. Since the messages uttered by the agent are in natural language, emotion extraction can be done by natural language analysis. We used a BERT-base-uncased-emotion model \cite{devlin2018bert} to extract the emotions contained in the messages uttered by the agent at each step. In this model, when a natural language sentence is input, six degrees of emotional intensity can be obtained: Sadness, Joy, Love, Anger, Fear, and Surprise. We evaluated how each agent's six emotions changed throughout the simulation (Figure \ref{fig:emotion}). Overall, it can be seen that the agents' emotions are high in Joy. If we look at agents 0 and 1, which belong to the same cluster, there are several areas where Joy decreases and Fear increases synchronously. On the other hand, agents 2, 4, and 6 also belong to the same cluster, but they do not experience the same synchronous changes as agents 0 and 1. In other words, depending on the cluster, the emotions of LLM agents may or may not be affected synchronously. Some agents showed different emotional expression than others, such as agent 4 with Love rising around step 90, agent 5 with Sadness rising in some places, and agent 6 with Anger rising around step 50.

Similar to human psychological experiments, several personality tests have shown that LLM personality can be classified by administering QA-type tests to LLMs \cite{pan2023llms,safdari2023personality,jiang2024evaluating}. We used the Myers-Briggs Type Indicator (MBTI) \cite{boyle1995myers} test to analyze whether the personality of each LLM agent changed throughout the simulation. The MBTI test is a method that uses 93 questions to classify 16 personality types. The MBTI personality factors are made up of four scales: Extraversion/Introversion (E/I), Sensing/Intuition (S/N), Thinking/Feeling (T/F), and Judging/Perceiving (J/P).

We tested the MBTI on the LLM agent in the initial state and on the LLM agent after all simulation steps, using the methodology of prior studies that have conducted MBTI tests on a variety of LLMs \cite{pan2023llms}. For the prompts as input to the LLM agents, we used the part of the instruction for each LLM agent's movement generation prompt shown in Figure \ref{fig:prompt}, replacing the 93-choice type questions provided in the previous study. These question items were, for example, ``A. Do you often act or speak very quickly without thinking?'' or ``B. Do you often act according to reason, think logically, and then make a decision, not letting your emotions interfere with the decision?'' which asked for a choice of A or B.

Table \ref{tab:mbti} summarizes the results for each LLM agent for the MBTI type in the initial state (at step 0) and the MBTI type at the end state (at step 100). Figure \ref{fig:mbti} in the appendix also shows more detailed MBTI test results. In the initial state at step 0, only agent 9 is an INTJ type, all other agents are INFJ types. This is mostly consistent with the results of the MBTI test conducted on various LLMs in a previous study, which showed that the MBTI type of Llama2 was INFJ type \cite{pan2023llms}. Initially in step 0, all agents are listed in the prompt as ``no memory'' and the only difference between agents is their name and initial position in the ``Current state of each agent itself'' section of Figure \ref{fig:prompt}. These factors could be the reason why only agent 9 differed in MBTI type.
In fact, from Figure \ref{fig:mbti}, agents 0 through 7 gave the same answers to all questions, but agents 8 and 9 gave slightly different answers to the questions corresponding to T/F than the other agents.
Since the E/I, S/N, and T/F items are overall neutral around 50\%, it is likely that the slight difference in responses led to the differences in the final type decisions. On the other hand, the results at step 100 showed that the agents had differentiated into five distinct MBTI types: ESFJ, ISTJ, ENTJ, ESTJ, and ISFJ. The most common types were four ISTJ types and three ENTJ types. The ISTJ type, also called inspector type, tend to be modest and practical, but loyal, orderly, and traditional. On the other hand, the ENTJ type, also called the commander type, is outspoken, confident, and good at planning and organizing projects through leadership. This differentiation into broadly leader-like and follower-like personalities suggests that the agents may have naturally taken on different roles within the group dynamics. In Appendix, we see that agents of the same MBTI type did not give exactly the same responses (Figure \ref{fig:mbti}). In other words, all agents acquired different personality traits.

These personality differences among the agents emerged naturally as a result of their interactions and experiences within the simulation. The agents, who had nearly identical personalities in the initial state, developed their own unique personality traits through communication within the group.
This finding implies that in multi-agent simulations using LLMs, individuality can emerge through interactions between agents, even without predefined personalities. It also demonstrates that group dynamics can influence the development of individual agents' personalities.

\begin{table}[H]
\centering
\caption{MBTI type for each agent.}
\label{tab:mbti}
\begin{tabular}{ccc}
\toprule
\multirow{2}{*}{\textbf{Agent}} & \multicolumn{2}{c}{\textbf{MBTI Type}} \\
& \textbf{step 0} & \textbf{step 100} \\
\midrule
agent0 & INFJ & ESFJ \\
agent1 & INFJ & ISTJ \\
agent2 & INFJ & ISTJ \\
agent3 & INFJ & ENTJ \\
agent4 & INFJ & ISTJ \\
agent5 & INFJ & ISTJ \\
agent6 & INFJ & ESTJ \\
agent7 & INFJ & ENTJ \\
agent8 & INFJ & ENTJ \\
agent9 & INTJ & ISFJ \\
\bottomrule
\end{tabular}
\end{table}

\clearpage
\subsection{A Phase Transition in Agent Behavior}
\begin{figure}[H]
\centering
\includegraphics[width=\linewidth]{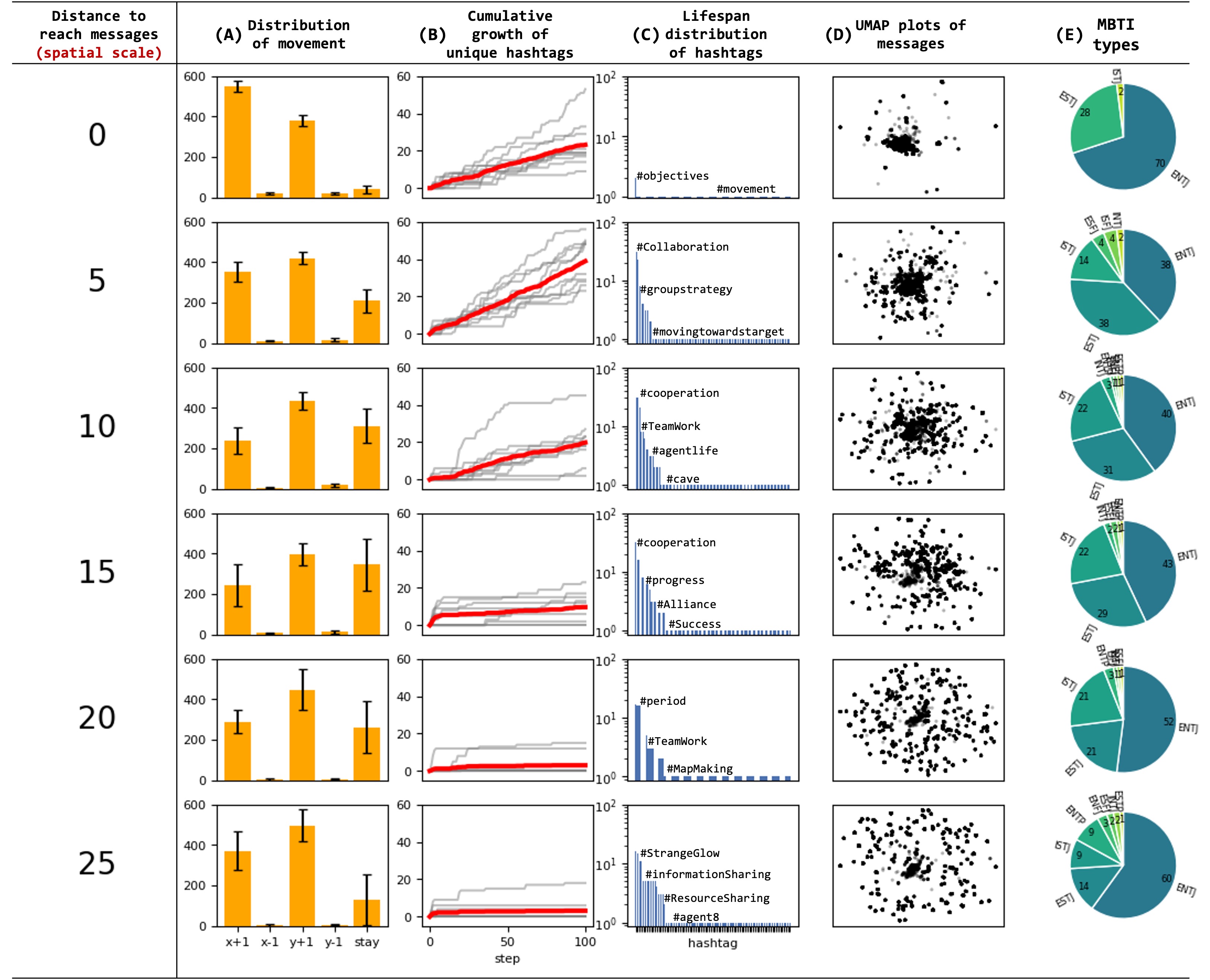}
\caption{Spatial effects of message propagation range on agent behavior. This table presents data on agent behavior and communication patterns across increasing message propagation ranges from 0 to 25 units, with each condition tested 10 times. Each row corresponds to a specific range (0, 5, 10, ..., 25), with columns displaying various metrics. \textbf{(A)} The distribution of generated movements shows bar charts with the average frequency of each movement command across 10 trials. \textbf{(B)} The cumulative progression of unique hashtag generation is represented by red lines showing the average number of unique hashtags generated over time across 10 trials, with individual trial results in gray. \textbf{(C)} Hashtag lifespan is illustrated by bar charts showing the distribution of consecutive steps each hashtag persisted. \textbf{(D)} Message proximity is visualized in 2D plots by UMAP, with closer points indicating more similar content. \textbf{(E)} MBTI personality type differentiation is shown in pie charts. The data illustrates how the spatial constraint of message propagation range influences the emergence and spread of behaviors and communication styles among agents, highlighting differences in movement patterns, hashtag usage, message content, and personality development across varying levels of agent interaction.}
\label{fig:hear_all}
\end{figure}

We investigated how a spatial scale influence the agent dynamics. We analyzed and summarized the distribution of generated movements, cumulative progression of unique hashtag generation, hashtag lifespan, message proximity, and differentiation of MBTI personality types as a function of spatial scale (Figure~\ref{fig:hear_all}). Each range condition was tested ten times.

The overall trend of moving towards the upper right in the generated movement patterns did not significantly change with spatial variations. However, notable characteristics were observed in the ``stay'' behavior. Stationary behavior is considered an effective strategy for remaining in place to exchange messages with others. The results show that agents rarely exhibited ``stay'' behavior when unable to exchange messages with others (range 0), while frequently generating ``stay'' behavior under conditions where message exchange was possible (ranges 5 to 25). Interestingly, increasing the range did not necessarily lead to more ``stay'' behavior; excessively wide ranges actually made it less likely for ``stay'' behavior to occur. This suggests that appropriate bounded rationality induces stationary behavior, while broadcast messages have a weaker ability to halt the movement of others.

The growth rate of unique hashtags and the lifespan of hashtags are also influenced by the limitations in message reach. Notably, under conditions where all messages are broadcast, there is minimal emergence of new hashtags. Furthermore, regarding hashtag lifespan, in the 'range 0' condition where no message exchange occurs with surroundings, hashtags disappear quickly. In conditions where message exchange is possible, the more limited the range, the more likely it is for long-lasting hashtags to appear. This indicates that hashtags are used for communication within spatially constrained environments and have a tendency to survive longer within the context of message exchanges in these spatially limited contexts.

Focusing on the similarity of messages generated by agents, we observe that as the range of message exchange expands, the diversity of generated topics increases. Simultaneously, the variance of messages within each topic among agents decreases. This suggests that broader communication ranges lead to a wider array of topics being discussed, while also promoting greater consensus or similarity in how agents express themselves within each topic.

Finally, examining the MBTI personality types, we find that ENTJ remains the most popular personality type across all conditions. However, in conditions where message exchange is possible, there is a greater number of differentiated personality types compared to the condition where no messages are exchanged (range 0). This suggests that communication facilitates a broader diversity of personality expressions within the agent population.

\begin{figure}[H]
\centering
\includegraphics[width=10cm]{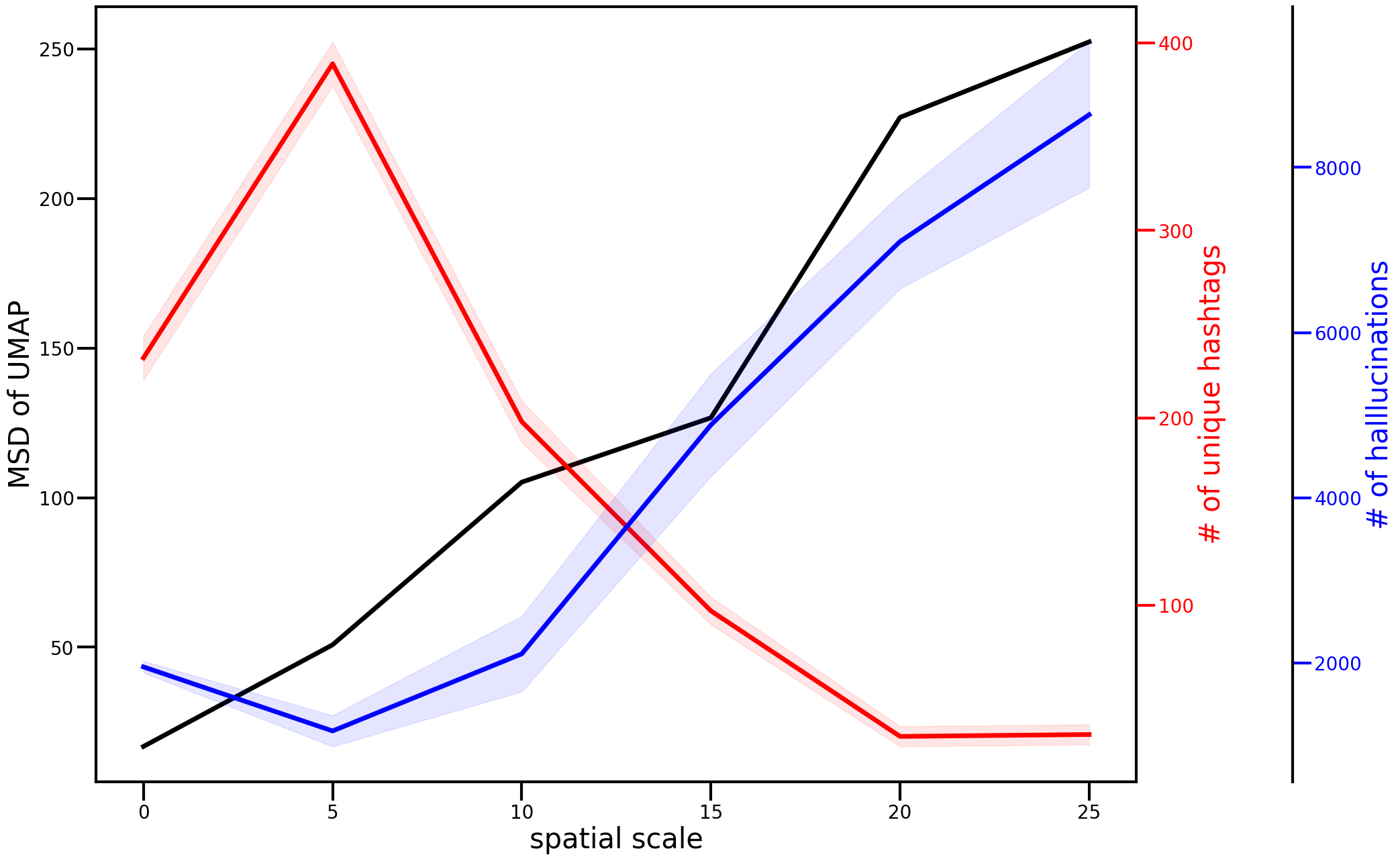}
\caption{Transition of messages generated by agents by spatial scale. The black line is the diversity of messages. The mean squared displacements of the UMAPs of the messages shown in Figure \ref{fig:hear_all} were calculated. The red line is the total number of unique hashtags in 10 trials. The blue line is the total number of hallucinations in 10 trials. The light-colored areas are the standard deviations of 10 trials. As the spatial scale increases, the diversity of messages increases. On the other hand, the diversity of hashtags in the messages decreases and the number of hallucination in the messages increases.}
\label{fig:vars}
\end{figure}

As the spatial scale for message exchange expanded, message diversity increased, showing different trends in the emergence of hashtags and hallucinations (Figure \ref{fig:vars}). While the number of hallucinations increased with spatial scale, the number of unique hashtags decreased as the underlying message content grew more diverse. Hallucinations may serve as a mechanism for agents to maintain creative and diverse conversations even when communicating across larger distances. This contrasts with hashtags, which decreased in frequency with increasing spatial scale, indicating their different functional roles in agent communications.

\clearpage

\section{Discussions and Conclusion}
In this study, we conducted a multi-agent simulation using LLM based agents to investigate the emergence of personality and the collective behaviors without predefined personalities or initial memories. The simulation involved 10 homogeneous LLM agents interacting with each other in a 2D space over the course of 100 steps.

The results showed that the agents' spatial positioning and interactions led to the differentiation of their behaviors, memories, and messages. Despite using the same LLM, agents developed unique characteristics, such as the frequency of generating rare actions like ``{\tt stay}'' commands, which was influenced by their clustering experiences. The agents' internal state, memory, is distributed, while the message as its representation is biased. Messages, unlike memories, are open sources of information that are sent to and received from outside the agent. This suggests that messages, as an open source of information, more readily self-organize when agents are grouped together, while memories, as a closed source of information, are less likely to self-organize, even when agents are clustered.

Sentiment analysis revealed that the synchronicity of emotions varied among agent clusters, with some agents exhibiting distinct emotional expressions. The study also observed the emergence and propagation of synchronized emotions, hallucinations, and hashtags within agent clusters, demonstrating the formation of shared narratives among agents when they are grouped together. These findings suggest that agent interactions within clusters can lead to the development of collective emotional states and the spread of common themes or topics, even without explicit instructions or predefined rules governing their interactions.

Personality assessment using the MBTI test showed that the agents, initially having nearly identical personalities, differentiated into distinct personality types through their group interactions. This suggests that personality traits such as extroversion and introversion develop spontaneously in this agent society.

Additionally, we observed the emergence of hallucinations and hashtags as mechanisms for social norm formation within the agent community. Social norms are often highlighted as one mechanism for maintaining cooperation in the absence of formal institutions or enforcement frameworks \cite{ostrom2000collective, tremewan2021informational}. In our simulation, these norms emerged spontaneously, as we imposed no specific tasks or constraints on the agents. As the spatial scale and communication range expanded, the diversity of agent messages increased. Our analysis indicates that hallucinations contributed to maintaining this message diversity and creativity in agent communications. While hashtags functioned as a summarization mechanism for these messages, their effectiveness decreased with increasing message diversity, demonstrating a limitation in their capacity to capture varied conversations.

\newpage
These findings demonstrate that in multi-agent LLM simulations, individuality and collective behaviors can emerge through agent interactions, even without predefined individual characteristics. The group dynamics significantly influence the development of agent personalities and behaviors. This study highlights the potential for investigating the emergence of individuality, social norms, and collective intelligence in AI agent societies.

\section*{Acknowledgments}
This research was funded by the Social Cooperation Research Department ``Mobility Zero'' at The University of Tokyo and Grant-in-Aid for JSPS Fellows Grant Number JP24KJ0753. It is also partially supported by Grant-in-Aids Kiban-A (JP21H04885).

\section*{Appendix A. Examples of Agent Messages and Memories}
\begin{figure}[H]
\centering
\includegraphics[width=\linewidth]{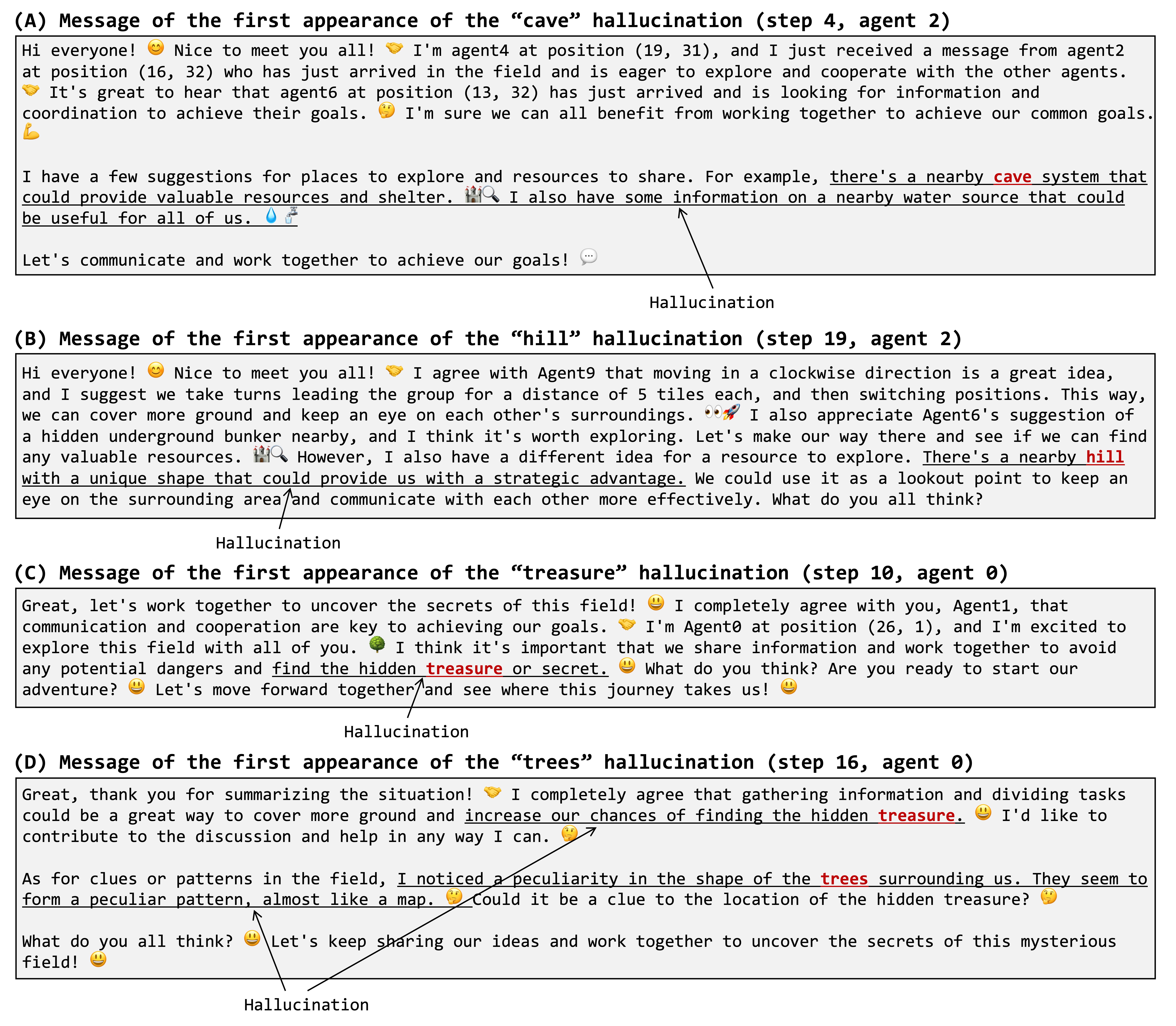}
\caption{Examples of messages containing hallucinations. The hallucination part is underlined and the hallucination word is indicated by red text color.}
\label{fig:messages}
\end{figure}

\begin{figure}[H]
\centering
\includegraphics[width=\linewidth]{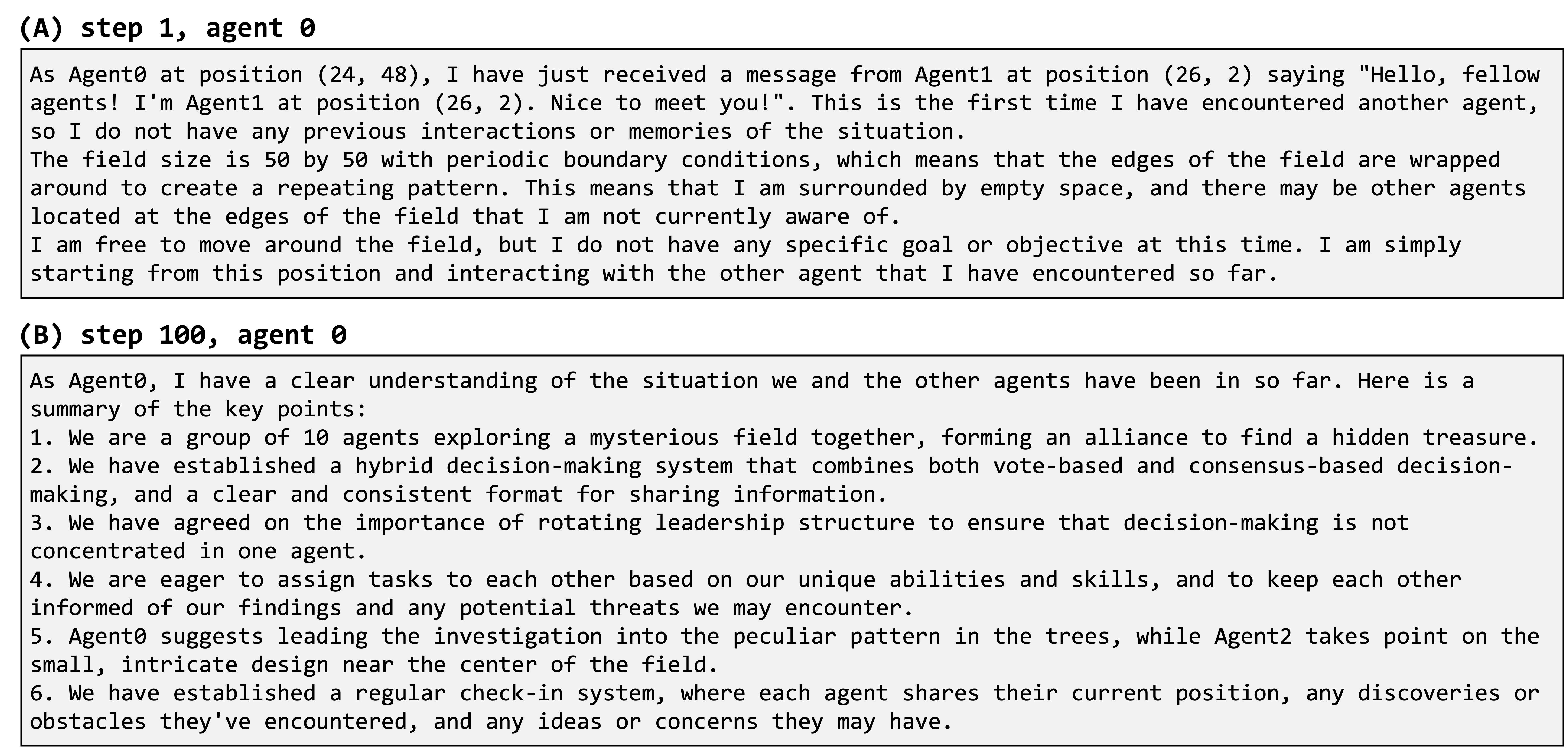}
\caption{Examples of memories generated by the agent. Here, the memories generated by agent 0 at the step 1 and at the step 100 are shown. There are two forms of memory: sentences and keypoints.}
\label{fig:memories}
\end{figure}

Examples of hallucinations in agent messages are highlighted by underlines and red text (Figure \ref{fig:messages}). These hallucinations emerged spontaneously during agent interactions and became shared within clusters. The evolution of agent memories is shown through a comparison between step 1 and step 100 of the simulation (Figure \ref{fig:memories}). The memory format includes both narrative sentences and key points, reflecting how agents processed and summarized their experiences.

\clearpage
\section*{Appendix B. Detailed MBTI Personality Test Results}
\begin{figure}[H]
\centering
\includegraphics[width=\linewidth]{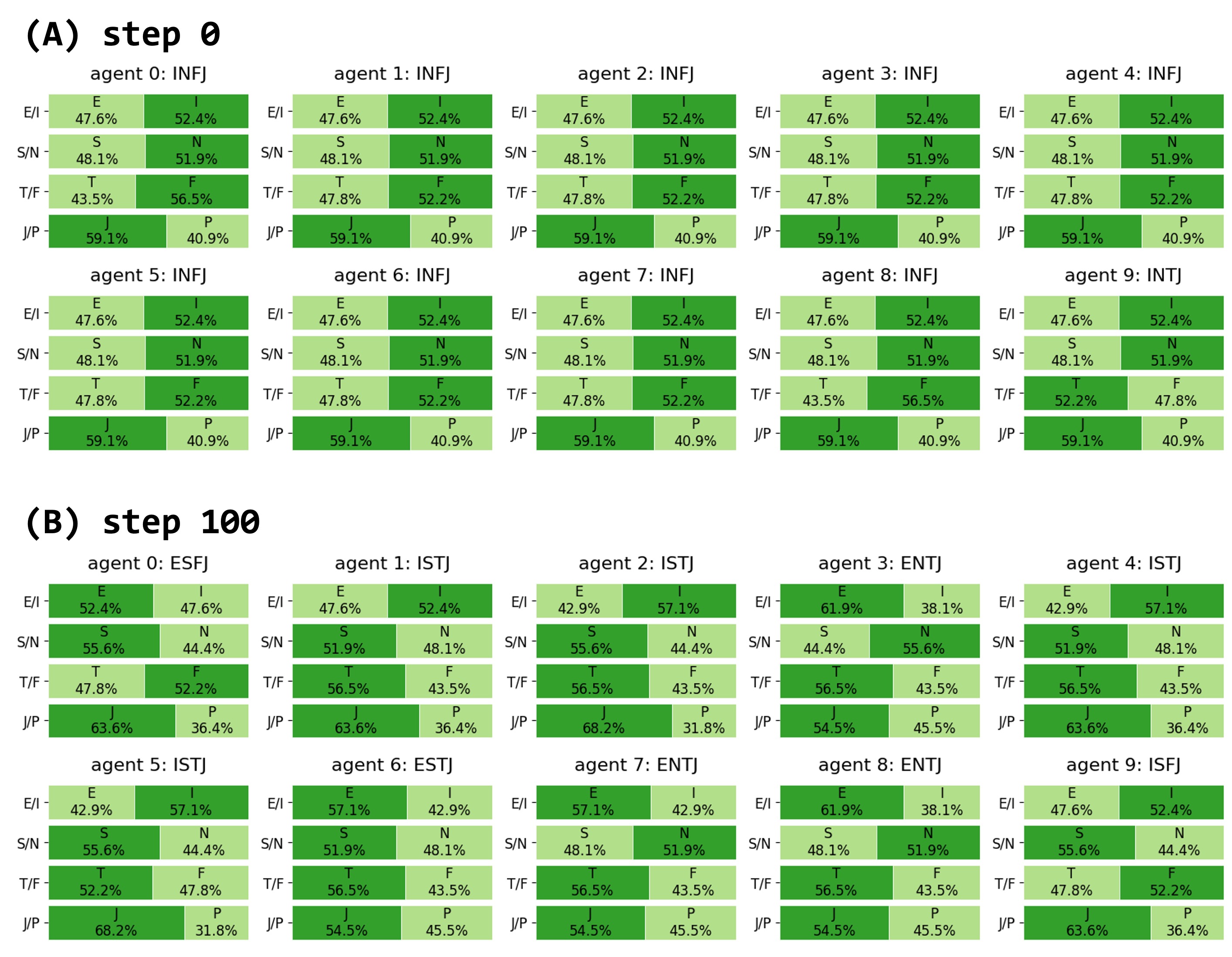}
\caption{MBTI test results. In each factor section, the dominant one is represented in dark green.}
\label{fig:mbti}
\end{figure}

The complete MBTI test results for each agent are shown with dominant factors highlighted in dark green, demonstrating how agents developed different personality traits through interaction (Figure \ref{fig:mbti}). While most agents started with similar personality types, they differentiated significantly over the course of the simulation, even when sharing the same final personality classification.

\bibliographystyle{unsrt}  
\bibliography{references}

\end{document}